\documentclass[10pt,conference,compsocconf]{IEEEtran}

\usepackage{booktabs}
\usepackage{paralist}
\usepackage{tikz}
\usepackage{pgf}
\usepackage{flushend}
\usepackage{url}
\usepackage{pgfplots}
\usepackage{graphicx}
\newcommand{\inw}[1]{\textit{#1}}  
\newcommand{\avmf}[1]{{\scshape#1}} 
\usetikzlibrary{arrows,automata}
\newcommand{\rel}[1]{\textsc{#1}}
\newcommand\Wocadi{Wocadi}
\newcommand{\quw}[1]{\textit{#1}}  
\newcommand{\trw}[1]{\textit{`#1'}}

\author{\IEEEauthorblockN{Tim vor der Br\"uck}
\IEEEauthorblockA{\textit{School of Computer Science and Information Technology} \\
\textit{Lucerne University of Applied Sciences and Arts (HSLU)}\\
Rotkreuz, Switzerland \\
tim.vorderbrueck@ffhs.ch}
\and
\IEEEauthorblockN{Marc Pouly}
\IEEEauthorblockA{\textit{School of Computer Science and Information Technology} \\
\textit{Lucerne University of Applied Sciences and Arts (HSLU)}\\
Rotkreuz, Switzerland\\
marc.pouly@hslu.ch}}

\title{\textbf{\Large Estimating Text Similarity based on Semantic Concept Embeddings}}

\begin{document}


\maketitle

\begin{abstract}
Due to their ease of use and high accuracy, Word2Vec (W2V) word embeddings enjoy  great success in the semantic representation of words, sentences, and whole documents as well as for semantic similarity estimation. However, they have the shortcoming that they are directly extracted from a surface representation, which does not adequately represent human thought processes and also performs poorly for highly ambiguous words.  Therefore, we propose Semantic Concept Embeddings (CE) based on the MultiNet Semantic Network (SN) formalism, which addresses both shortcomings. The evaluation on a marketing target group distribution task showed that the accuracy of predicted target groups can be increased by combining traditional word embeddings with semantic CEs. 
\end{abstract}

\begin{IEEEkeywords}
Concepts, MultiNet, concept embeddings, semantic similarity estimation.
\end{IEEEkeywords}

\section{Introduction}
Word2Vec (W2V) word embeddings became a popular method for creating semantic representations of words, sentences, and whole documents as well as their semantic similarity estimation. Their  key success factors are ease of use,
scalability, and performance. However, there are also important downsides. Consider, for example, a word just in front of an embedded sub-clause. Potentially, the words just succeeding this sub-clause might constitute a more characteristic word context than the words actually occurring inside this sub-clause despite being farther apart. So, in some situations, considering only the word context based on the surface structure may not  be the optimal choice. Hence, we hereby introduce semantic Concept Embeddings (CEs) that consider the neighborhood in a semantic network (SN). We expect a meaning-oriented structure such as an SN to provide  a much more adequate  description of neighborhood since it is oriented on human thought processes.
The SNs are automatically constructed from arbitrary surface texts by the \Wocadi\ parser \cite{hartrumpf02}, which combines manual word function-oriented rules with statistical disambiguation methods. In a pre-study, \Wocadi\ obtained superior results than the only state-of-the-art semantic role labeling parser supporting German out-of-the-box, which is AMR-Eager-Multilingual \cite{Damonte_etal17} (AMR stands for Abstract Meaning Representation), a Deep Learning based approach.  The SNs as generated by Wocadi are much more comprehensive than, for example, WordNet \cite{fellbaum98} or GermaNet \cite{hamp97}, since they can represent the semantics of arbitrary texts and not only ontologies. 

The CEs obtained from these SNs are then employed in the task of assigning participants of an online contest to certain marketing target groups, called youth milieus, by analyzing short text snippets provided by these participants.


Finally, we applied our estimate to the scenario of distributing participants of an online contest into several target groups called youth milieus by exploiting short text snippets they were asked to provide. 
The evaluation revealed that our novel estimators performed superior to several baseline methods for this scenario.

The remainder of the paper is organized as follows. In the next section, we look into several state-of-the-art methods for estimating semantic similarity. 
Sect.~\ref{sec:networks}  introduces the MultiNet semantic network formalism. 
In Sect.~\ref{sec:semantic_similarity}, we describe in detail how these networks can be used for estimating semantic similarity between two texts.
Our  application scenario for our proposed semantic similarity estimates  is given in 
Sect.~\ref{sec:market_segmentation2}. Sect.~\ref{sec:evaluation} describes the conducted evaluation, in which we compare our approach with several baseline methods. The results of the evaluation are discussed in Sect.~\ref{sec:discussion}.   
Finally, this paper concludes with Sect.~\ref{sec:conclusion}, which summarizes the obtained results.  

\section{Related Work}


The main application area of knowledge graph embeddings is link prediction \cite{tourillon_etal16}.
They are very rarely used to estimate the semantic similarity of texts, confer as an example for the latter the approach of Goikoetxea et al. \cite{goikoetxea16}. 
 They generate random walks on WordNet to extract sequences of words, where the lexicalization is randomly chosen for the associated synset nodes. These sequences are then fed into the ordinary W2V to create (ontology) embedding vectors. They evaluated several possibilities to combine such vectors with ordinary word embeddings obtained from large corpora like  averaging or concatenating them. Another embeddings extraction method for WordNet synsets is proposed by Kutuzov et al. \cite{kutuzov18}. They obtained superior results to random walks by obtaining the embedding vectors as a result of an optimization problem that ensures local (graph neighbor) and global (user annotations) consistency.  Note that WordNet is much smaller than Wikipedia, which we use as a basis for our approach. Furthermore, MultiNet concepts have some advantages over MultiNet synsets, since in some cases word senses  cannot be fully captured by synsets.

An approach to obtain CEs from ordinary surface texts is proposed by Mencia et al. \cite{mencia-etal-2016-medical}. In particular,  they generate  CE
by counting co-occurrences of  \emph{concepts} akin  to obtaining GloVe \cite{pennington_etal14} word embeddings.  A similar approach was introduced by Beam et al. \cite{beam2018clinical} that directly  uses  GloVe to obtain the semantic concept vectors.
However, both approaches do not employ a proper Word Sense Disambiguation to obtain their concept representation. Instead,  they restrict their embeddings vocabulary to certain medical terms, which are usually specific enough to represent only  unique word senses.


After their extraction, word or concept embeddings can be used to estimate document similarity as follows:
\begin{compactenum}
\item The  embeddings (often weighted by the tf-idf coefficients of the associated words \cite{brokos16}) are looked up in a hashtable for all the words in the two documents to compare. These embeddings are determined beforehand on a very large corpus typically using either the skip-gram or the continuous bag of words variant of the W2V model \cite{mikolov13}.\ The skip-gram method aims to predict the textual surroundings of a given word by means of an artificial neural network.\ The influential weights of the one-hot-encoded input word to the nodes of the hidden layer constitute the embedding vector. For the so-called \emph{continuous bag of words} method, it is just the opposite, i.e., the center word is predicted by the words in its surroundings.
\item The centroid over all word embeddings belonging to the same document is calculated to obtain its vector representation.
\end{compactenum}


Alternatives to  W2V are  {GloVe} \cite{pennington_etal14}, which is based on aggregated global word co-occurrence statistics and the  Explicit Semantic Analysis (or shortly ESA) \cite{markovitch09}, in which  each word is represented by the column vector in the tf-idf matrix over Wikipedia.

The idea of W2V can be transferred to the level of sentences as well.\ In particular, the so-called Skip Thought Vector (STV) model \cite{kiros15} derives a vector representation of the current sentence by predicting the surrounding sentences.


If vector representations of the two documents to compare were successfully established, a similarity estimate can  be obtained by applying the cosine measure to the two vectors. \cite{song_roth15} propose an alternative approach for ESA word embeddings that establishes a bipartite graph consisting of the best matching vector components by solving a linear  optimization problem. The similarity estimate for the documents is then given by the global optimum of the objective function. However, this method is only useful for sparse vector representations. In the case of dense vectors, \cite{mijangos15} suggests applying the Frobenius kernel to the embedding matrices, which contain the embedding vectors for all document components (usually either sentences or words, cf. also  \cite{hong_etal15}).\ 
However, crucial limitations are that the Frobenius kernel is only applicable if the number of words (sentences respectively) in the compared documents coincide   and that a word from the first document is only compared with its counterpart from the second document. Thus, an optimal matching has to be established already beforehand.

Word embeddings as described so far are represented by a fixed word-vector mapping, which means that  these word vectors do not vary depending on the current word context.  This is different for Elmo \cite{Peters_et_al_2018} and Bert \cite{devlin2019,alsentzer2019publicly} embeddings that take this context into account basically realizing a kind of word sense disambiguation.

Before going into more details of our method, we first want to introduce the MultiNet SN formalism.

\section{SNs based on the MultiNet formalism}
\label{sec:networks}

In contrast to other popular knowledge graphs or SNs like WordNet \cite{fellbaum98}, OdeNet \cite{Siegel2021OdeNetCA} or Yago \cite{yago15} that represent ontologies,
MultiNet is designed to grasp the entire semantics of natural language. Therefore, MultiNet embeddings can be trained on much larger data sets than the formerly mentioned knowledge graphs, in case of this paper on the entire German Wikipedia.
SNs of the MultiNet (Multilayered Extended SNs) formalism \cite{helbig06l} allow to homogeneously represent
the semantics of single words, phrases, sentences, texts, or text collections.

An SN node represents a concept, while an SN arc expresses a relation between
two concepts. A concept \emph{lemma.x.y} is represented by a lemma, and two numbers, where the first (x)
denotes the homograph and the second (y) the polyseme.
In addition, each node is semantically classified by a \inw{sort} from a hierarchy of 45 sorts organized in a taxonomy.

Furthermore, a node has an inner structure (depending on its sort) containing
\inw{layer features} like \avmf{card} (cardinality) and \avmf{refer}
(referential determinacy).

The \Wocadi\ parser
can construct SNs of the MultiNet formalism 
for German phrases, sentences, or texts.
During this process, SNs and syntactic dependency structures are built.

An important component of this deep syntactico-semantic analysis of natural language 
is HaGenLex, a semantically based computer lexicon \cite{hartrumpf_helbig_etal03}.
This lexicon not only lists verb valencies, but also their syntactic and
semantic types.
Consider, for example, the German verb \quw{essen} (\trw{eat}).
Sentences like \quw{Die Birne isst den Apfel.} (\trw{The pear eats the apple.})
are rejected because semantic selectional restrictions are violated.
Besides this comprehensive lexicon with around 28,000 entries,
 a shallow lexicon, many name lexicons, and a sophisticated compound analysis is applied 
to achieve the parser coverage required for natural language processing applications.

Disambiguation is realized by specialized modules which work
with symbolic rules and disambiguation statistics derived from
annotated corpora. Currently, such modules exist for
(intrasentential and intersentential) coreference resolution,
the attachment of prepositional phrases,
and the interpretation of prepositional phrases.

The following MultiNet relations and functions are relevant to this paper \cite{helbig06l}:
\begin{compactitem}
\item \rel{agt}: Definition: In its standard interpretation, the expression (e \rel{agt} o)  characterizes the relation between an event e or, to be specific, an action e and a conceptual object o which actively causes e (i.e. o is originating/sustaining/giving rise to e). In other words, o is the active object
(the agent or carrier of the action).
\item \rel{chea}: The statement (e \rel{chea} a) expresses the connection between an
event e  (usually a verb) and an abstract concept a (usually a noun) which agree, at least partially, in their meaning. Concepts connected by \rel{chea} correspond to each other in a
systematic way.  Example: (inform.1.1 \rel{chea} information.1.1)
\item \rel{chpa}: The statement (p \rel{chpa} c) establishes a connection between a
property p (usually adjective) and an abstract concept (usually a noun) c which
is semantically close to p and whose meaning is systematically related to p. Example:
(cold.1.1 \rel{chpa} coldness.1.1). 
\item \rel{sub}: The expression ($o_1$ \rel{sub} $o_2$) specifies that the individual or generic
concept $o_1$ is subordinate (a hyponym) to the generic concept $o_2$, i.e. everything derivable
for $o_2$ is also valid for $o_1$.
\end{compactitem}

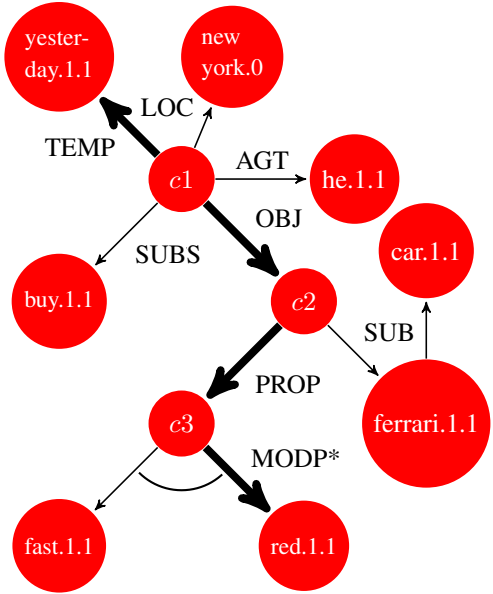
\begin{figure}
\centering
  \begin{tikzpicture}[>=stealth',shorten >=1pt,auto,node distance=2.3cm,
                    semithick,scale=0.1]
  \tikzstyle{every state}=[fill=red,draw=none,text=white]

  \node[state] (A)                    {$c1$};
 \node[state]         (B) [right of=A] {he.1.1}; 
  \node[state]         (P) [below right of=A] {$c2$};
  \node[state]         (E) [below left of=A] {\small buy.1.1};
  \node[state]         (C) [below left of=P]       {$c3$};
  \node[state]         (D) [below right of=P]   {ferrari.1.1} ;
  \node[state]         (F) [below left of=C]  {\small fast.1.1};
  \node[state]         (G) [below right of=C] {\small red.1.1} ;
  \node[state]         (H) [above left of=A] {$c4$};
  \node[state,align=left]         (I) [right of=H]{\small new\\\small york.0};
  \node[state,align=left]         (J) [above left of=A] {\small yester-\\\small day.1.1};
  \node[state]         (K) [above of=D] {car.1.1};

\draw [->] (A) edge node {AGT} (B);
\draw [->] (A) edge node {SUBS} (E);
\draw [->, line width=3pt] (A) edge node {OBJ} (P);
\draw [->] (A) edge node {LOC} (I);
\draw [->,line width=3pt] (A) edge node {TEMP} (J);
\draw [->, line width=3pt] (P) edge node {PROP} (C);
\draw [->] (P) edge node {SUB} (D);
\draw [->] (D) edge node {} (K); 
\draw [->, line width=3pt] (C) edge node {MODP*} (G);
\draw [->] (C) edge node {} (F);

\draw [thick,domain=225:315] plot ({8*cos(\x)}, {-33.4+8*sin(\x)});
\draw (0.8,-9.4) -- (1,-9.1);
\draw (0.7,-9.15) -- (1,-9.15);

\end{tikzpicture}
  \caption{Example for a random walk (yesterday.1.1 \rel{temp}$^{-1}$  \rel{obj}  \rel{prop}  \rel{modp*} red.1.1) in the SN.}
  \label{lab:random_walk}
\end{figure}

An example for an SN following the MultiNet formalism is given in Figure~\ref{lab:random_walk}.

\section{Using semantic CEs for estimating semantic similarity}
\label{sec:semantic_similarity}

\subsection{Procedure}

To better understand why a representation as an SN can be beneficial, let us consider the following example sentence:
\emph{The car he bought yesterday in New York is a fast red Ferrari.}
This sentence contains the embedded relative clause \emph{he bought yesterday in New York}. 
Actually, the noun phrase \emph{fast red Ferrari} is much more characteristic for the word  car 
than the subclause \emph{he bought yesterday in New York} despite being farther away from that word.
The representation of this sentence as an SN on MultiNet is shown in Figure~\ref{lab:random_walk}.
The main node representing the entire sentence is c1, which is a specific buying operation expressed by the relation \rel{subs}. The relation \rel{loc} specifies the location of the event, in this case \emph{New York}. The trailing .0 at the concept name denotes the fact that the concept is identified by a proper name. The agent (relation: \rel{agt}) of the event is \emph{he.1.1}, which can potentially be resolved by an anaphora resolution and replaced by its antecedent. The Ferrari that is bought is identified by the concept c2,  connected to c1 by an object (\rel{obj}) relation. 
The red Ferrari is also fast, which is expressed by a modifier (\rel{modp*}) combined with a property relation (\rel{prop}).

Now let us consider a second example given by the following two sentences:
\emph{Pete kauft einen neuen roten Ferrari.}
(\emph{Pete buys a new red Ferrari.})
 und
\emph{Ein neuer roter Ferrari wird von Pete gekauft.}
(\emph{A new red Ferrari is bought by Pete.})
Both sentences express the same meaning 
but the surface structure is different. The representation in passive voice
contains the additional word (\emph{wird} (English: \emph{was})), which is in German a function word that can also express future tense, i.e., this word is ambiguous and
can therefore add additional noise to the word vectors. The SNs of both sentences however coincide, which results in more exact embedding vectors.

Let us consider as a third example the word  \emph{space}, which can either denote a certain geographical area or the universe. Actually, the word vector of \emph{space} integrates all possible word senses. Thus, the cosine distance between the word vectors of \emph{space}  and \emph{planet} should be  smaller than for \emph{universe}  and \emph{planet}. 
However, if it is known from the word context that \emph{space} means \emph{universe} in a certain text, then  \emph{space} and \emph{planet} should be similarily semantically related as  \emph{universe} and \emph{planet}. Using our SN-based approach, we no longer generate vectors for surface words but for word senses instead, which solves this issue.
 A concrete German example comprises the two German words \emph{Lok} (locomotive) and \emph{Zug} (move(ment) / train)  and  is given in Table~\ref{tab:example}. It demonstrates that the concept \emph{lok.1.1} (locomotive.1.1) has a higher cosine similarity to the word sense of \emph{Zug}  denoting \emph{train} (\emph{zug.1.1}) than to the word sense \emph{zug.1.2} denoting move(ment), and this cosine similarity also exceeds the one  between the word vectors of \emph{Zug} and \emph{Lok}. 
\begin{table}
 \caption{Cosine measure applied to word and concept vectors.}
  \label{tab:example}
        {\small
          \begin{tabular}{p{2cm}p{2cm}p{2cm}l}
    \toprule
    Wort    1 / & Word 2 / & Example for  & Cosine   \\
    Concept 1   & Concept 2&  Concept 2         &          \\
    \midrule
    Lok (locomotive)      & Zug (move(ment)/train)  & &0.428 \\
    lok.1.1   & zug.1.1      & Der Zug nach London f\"ahrt ab. (The train to London is departing.) & 0.633 \\
    lok.1.1                & zug.1.2      & Zug der V\"ogel (bird migration (movement)) & 0.321 \\
    \bottomrule
  \end{tabular}
  }
 
\end{table}



Our workflow to create semantic CEs is as follows:
\begin{compactitem}
\item Parse the German Wikipedia  by means of \Wocadi\ to create SNs
\item Create random walks on these SNs
\item Feed the random walks into W2V to create CEs
\end{compactitem}
Note that  we obtained already a fully parsed Wikipedia from Sempria GmbH, a spin-off of the
Distance University of Hagen.
Once the CEs are generated, they can be exploited to estimate semantic
text similarity in the following way:

\begin{compactitem}
\item Apply a Word Sense Disambiguation to both texts to obtain MultiNet concepts.
\item Create CEs for both concept representations.
\item Determine the two CEs centroids.
\item Estimate the semantic similarity between two texts using the cosine measure.
\end{compactitem}

There are several approaches to obtaining embeddings from knowledge graphs and SN like RotatE \cite{rotate19}, TransE \cite{bordes_etal13}, or TorusE \cite{toruse}.
We decided to train W2V on random walks over the MultiNet-SNs, which has the advantage that hereby word embeddings and CEs are better
comparable since both are obtained using the same method, namely W2V.
To generate such a random walk, one first picks an initial node from an SN randomly.
Afterwards, one randomly chooses a neighboring node and repeats this process using the newly picked node.
Additionally, one generates in each step a continuous random number between 0 and 1 and terminates the  random walk, if this number assumes a value below a certain (small) threshold.
Note that in contrast to a path, edges can appear several times in a certain random walk and one can also move back and forth on the same edge. 
The random walk representation is then 
given by an interleaving sequence of nodes and edges.
If an edge is followed against the arc direction, we note down the inverse  of the relation associated with the arc.
Let us consider as an example the SN shown in Figure~\ref{lab:random_walk}, in which the following  possible random
walk is already highlighted:\\
yesterday.1.1 \rel{temp}$^{-1}$ c1 \rel{obj} c2 \rel{prop} c3 \rel{modp*} red.1.1 \\ 
Since the  inner nodes like c1 and c2 are named arbitrarily without
expressing any sort of meaning, we remove such nodes from the random walk.
Thus, our modified random walk becomes: \\ 
yesterday.1.1 \rel{temp}$^{-1}$  \rel{obj}  \rel{prop}  \rel{modp*} red.1.1 \\ 
In case of symmetric relations $\mathit{Rel}$ with $\mathit{Rel}=\mathit{Rel}^{-1}$, we only consider 
the non-inverted relations in the random walk. In our example, the random walk does not contain any  symmetric relations so it stays unchanged.

To estimate the semantic similarity between two texts using CEs, one first has to map the words appearing in these texts to concepts,
which boils down to applying a Word Sense Disambiguation. 
One possibility to accomplish this task is to parse both texts using \Wocadi, which generates for each text an SN and also
establishes a mapping from each word to the associated concept. 
However, note that \Wocadi\ is not freely available but requires a license from Sempria GmbH.
Therefore, we use a different approach based on so-called word-concept embeddings. 
As the name insinuates, word-concept embeddings denote concept vectors, which are based on a surface oriented approach.
In particular, we obtain a word-concept vector of some concept $\mathit{c}$ by averaging the word vector centroids of all sentences, where c occurs in the associated SN. We then chose the word sense c of a word w occurring in a certain sentence s, where the word-concept vector of $\mathit{c}$ is most similar to the centroid of $\mathit{s}$.

\section{Application Scenario: Market Segmentation}
\label{sec:market_segmentation2}

Market segmentation is one of the key tasks of a marketer. Usually, it is accomplished by clustering over behaviors as well as demographic, geographic, and psychographic variables  \cite{lynn11}. In this paper, we will describe an alternative approach based on unsupervised natural language processing. 
In particular, our business partner operates a commercial youth platform for the Swiss market, where registered members get access to third-party offers such as discounts and special events like concerts or castings.\ Actually, several hundred online contests per year are launched over this platform sponsored by other firms, an increasing number of them
 require the members to write short free-text snippets, e.g.\ to elaborate on a perfect holiday at a destination of their choice in case of a contest sponsored by a travel agency.\ Based on the results of a broad survey, the platform provider's marketers assume five different target groups (called \emph{youth milieus}) being present among the platform members:\ \emph{progressive postmodern youth} (people primarily interested in culture and arts), \emph{young performers} (people striving for a high salary with a strong affinity to luxury goods), \emph{freestyle action sportsmen}, \emph{hedonists} (rather poorly educated people who enjoy partying and disco music) and \emph{conservative youth} (traditional people with a strong concern for security).\ A sixth milieu called \emph{special groups} comprises all those who cannot be assigned to one of the upper five milieus.\ For each milieu (with the exception of \emph{special groups}) a keyword list was manually created by describing its main characteristics.\ For triggering marketing campaigns by marketers, an algorithm shall be developed that automatically assigns each contest answer to the most likely target group:\ we propose the youth milieu as best match for a contest answer, for which the estimated semantic similarity between the associated keyword list and user answer is maximal.\ In case the highest similarity estimate falls below the 10 percent quantile for the distribution of highest estimates, the special groups milieu is selected (cf. also \cite{vorderbrueck_pouly19}).

Since the keyword list typically consists of nouns (in the German language capitalized) and the user contest answers might contain a lot of adjectives and verbs as well, which do not match very well to nouns in the W2V  vector  representation, we actually conduct two comparisons for our W2V based measures, one with the unchanged user contest answers and one by capitalizing every word beforehand. The final similarity estimate is then given as the maximum value of both individual estimates.

The same procedure does not work with CEs, since concepts are always written lower case in MultiNet. To identify corresponding nouns for adjectives and verbs, we use the MultiNet relations \rel{chea} and \rel{chpea}. For instance, (inform.1.1 \rel{chea} information.1.1) or
(cold.1.1 \rel{chpa} coldness.1.1). These lexical relations are directly stored in the semantic lexicon named HaGenLex. Note that for better illustration, we gave English examples although we investigated German texts.


\section{Evaluation}
\label{sec:evaluation}


For evaluation of the marketing group segmentation task, we selected three online contests (language:\ German), where people elaborated on their favorite travel destination (contest 1), speculated about potential experiences with a pair of fancy sneakers (contest 2), and explained why they emotionally prefer a certain product out of four available candidates.\ To provide a gold standard, three professional marketers from different youth marketing companies annotated independently the best matching youth milieus for every contest answer.\ We determined for each annotator individually his/her average inter-annotator agreement with the others (Cohen's kappa). The minimum and maximum of these average agreement values are given  in Table~\ref{tab:kappa}.\ Since for contest 2 and contest 3, some of the annotators annotated only the first 50 entries (last 50 entries respectively), we specified min/max average kappa values for both parts. We further compared the youth milieus proposed by our unsupervised matching algorithm with the majority votes over the human experts' answers (see~Table~\ref{tab:accuracy}) and computed its average inter-annotator agreement with the human annotators (see again Table~\ref{tab:kappa}). 


\begin{table}
\centering
\caption{Minimum and maximum average inter-annotator agreements (Cohen's kappa).}
\label{tab:kappa}
{\small
\begin{tabular}{lrrr}
\toprule
Method & \multicolumn{3}{c}{Contest} \\
        & 1 & 2 & 3 \\
\midrule
Min kappa  &0.123  &  0.295/0.030& 0.110/0.101\\
Max. kappa &0.178  &  0.345/0.149& 0.114/0.209\\
\midrule
\midrule
\# Entries & 1544 & 100 & 100 \\
\bottomrule
\end{tabular}
}
\end{table}

\begin{table}
  \centering
\caption{Acc. values for several estimates. RW: method of \cite{goikoetxea16} applied to OdeNet, CE=Semantic Concept Embeddings, STV=Skip Thought Vectors, CE+W2V: weighted averaging cosine over word and CEs.} 
\label{tab:accuracy}
{\small
\begin{tabular}{lrrrr}
\toprule
Method&1&2&3&Total\\
\midrule
Random&0.152&0.090&0.167&0.146\\
Jaccard&0.150&0.194&0.045&0.142\\
GloVe&0.203&0.254&0.303&0.222\\
RW&0.294&0.254&0.227&0.281\\
W2V&0.299&$\mathbf{0.313}$&0.258&0.296\\
STV&0.241&0.194&$\mathbf{0.348}$&0.249\\
Elmo&0.150&0.224&0.258&0.173\\
Bert&0.221&0.209&0.212&0.218\\
CE&0.274&0.299&0.258&0.275\\
CE+W2V&$\mathbf{0.305}$&0.299&0.303&$\mathbf{0.304}$\\
\bottomrule
\end{tabular}

}
\end{table}


\begin{figure}
  \centering
\includegraphics[width=0.31\textwidth]{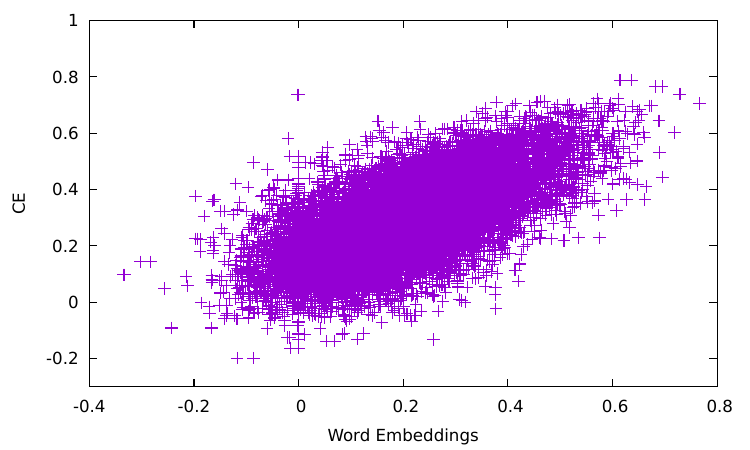}
\caption{Scatterplot between similarity estimate based on word embeddings (x-axis) and semantic CEs (y-axis).}
\label{sec:scatterplot}
\end{figure}

We evaluated both, our similarity estimate based on CEs alone  and its weighted mean with W2V (weight for CE: 0.2, weight for W2V: 0.8).

The W2V, CEs, and  Skip Thought Vectors (STV)  were trained on the German Wikipedia. 
The actual document similarity estimation for STVs is accomplished by the usual centroid approach. An issue we are faced with for our  evaluation scenario of market segmentation (see Sect.~\ref{sec:market_segmentation2}) is that STVs are not  bag-of-word models but actually take the sequence of the words into account and therefore the obtained similarity estimate between the milieu keyword list and contest answer  would  depend on the keyword ordering. However, this order could  have  arbitrarily been chosen by the marketers and might be completely random. A possible solution is to compare the contest answers with  all possible permutations of keywords and determine the maximum value over all those comparisons. However, such an approach would be infeasible already for medium keyword list sizes. Therefore, we apply  for this scenario a beam search to extend the keyword list iteratively while  keeping only the n-best performing permutations.

We also applied the approach of Goikoetxea et al. \cite{goikoetxea16}, another embeddings approach based on a linguistic network. Originally, the embeddings have been created by applying random walks on WordNet choosing randomly  lexicalizations for the synset nodes. Since WordNet is not available for German and we failed to obtain an academic license for the largest 
German-based linguistic network GermaNet, we applied their approach to the freely available German linguistic ontology OdeNet \cite{Siegel2021OdeNetCA}. Goikoetxea et al. proposed several combination methods of ordinary W2V and their linguistic network-based word vectors, where the concatenation of both  performed best. This is also the approach we evaluated in this paper. 

Note that we did not conduct a hyperparameter search on our input parameters like the random walk sentence restart threshold or the 10\% quantil for  selecting the Special Groups milieu. Instead we selected values based on experience that gave good results in practice. The window size for the embeddings has been chosen with seven rather large to obtain a reasonable number of concepts in the window.

\label{sec:discussion}

Consider  the annotations provided by the marketing experts, the evaluation showed that the inter-annotator agreement values vary  strongly for contest 2 part 2 (minimum average annotator agreement according to  Cohen's kappa of 0.03 while the maximum is 0.149, see~Table~\ref{tab:kappa}). In general, the kappa values, which estimate inter-annotator agreement  are rather small, which insinuates that the manual labeling  of the  contests and therefore also the automatic labeling process
proved to be quite challenging. 

In combination with word embeddings, semantic CEs  outperforms on our three contests word embeddings alone as well as the approach of Goikoetxea et al. as well as several deep learning based approaches like German Bert model, multilingual Elmo model, and Skip Thought Vectors. Still, semantic CEs alone perform currently worse than ordinary word embeddings. Sources of errors are the  word sense disambiguation (especially the word sense disambiguation of Wocadi) and the lemmatizer LemmaGen, which is currently trained on Wiktionary which provides a good coverage of generic nouns but falls short in terms of named entities. Additionally,  \Wocadi\ proved to be quite vulnerable in case of misspelled words or highly complex syntactic structures. In such cases it often only produces a chunk part
or the parsing process fails altogether. 
Figure~\ref{sec:scatterplot} shows a scatterplot, where we compare the similarity estimate based on word embeddings (x-axis) with its counterpart based on CEs (y-axis). The figure shows that there is a considerable correlation between both estimates.
One advantage of our method over using Bert and Elmo embeddings is that the marketer can directly specify the intended word senses as key words for the youth milieus. This is especially important if the keyword list is small and contains not enough textual contexts for Bert and Elmo
embeddings to perform well.


\section{Conclusion and Further Work}
\label{sec:conclusion}
We proposed a novel semantic similarity measure based on semantic CEs.
These embeddings were  extracted from an SN using random walks on them.
All the SNs were automatically generated from Wikipedia using the
\Wocadi\ parser. 
The evaluation showed that semantic CEs outperformed W2V embeddings when
used in combination with the latter on the task of assigning participants to the best matching marketing target group. 
There is still space for further improvement.
For instance,  inner nodes are currently eliminated from the random walk and therefore not used at all. Instead one could replace them with their associated ontological sort (see Section~\ref{sec:networks}).

Currently, our system depends on the \Wocadi\ parser, which is not freely available. Therefore, we aim to test, whether similar results can be obtained, if one uses freely available semantic role labeling parsers instead of \Wocadi. A concept representation of each word could hereby be obtained by determining the  WordNet / OdeNet / GermNet synset group that best fits to the given context as obtained by a Word Sense Disambiguation.

\section*{Acknowledgement}
Hereby we want to thank Sven Hartrumpf and Sempria GmbH for providing us with semantic network parses from  the entire Wikipedia.
\bibliographystyle{IEEEtran}
\bibliography{semantic_concept_embeddings}

\end{document}